%% file: main.tex
\documentclass[10pt,twocolumn,letterpaper]{article}

\usepackage[pagenumbers]{cvpr} 
\usepackage{multirow}
\usepackage{algorithm}
\usepackage{algorithmic}

\usepackage{url}

\definecolor{cvprblue}{rgb}{0.21,0.49,0.74}
\usepackage[pagebackref,breaklinks,colorlinks,allcolors=cvprblue]{hyperref}

\def\paperID{} 
\def\confName{CVPR}
\def\confYear{2026}

\title{FiDeSR: High-Fidelity and Detail-Preserving One-Step Diffusion Super-Resolution}

\author{
Aro Kim$^{1*}$ \quad
Myeongjin Jang$^{1*}$ \quad
Chaewon Moon$^{1}$ \quad
Youngjin Shin$^{1}$ \quad
Jinwoo Jeong$^{2}$ \quad
Sang-hyo Park$^{1\dagger}$ \\
$^{1}$Kyungpook National University \quad $^{2}$Korea Electronics Technology Institute \\
{\small \{arokim37, jmj4431, moonchaewon2, 2021112241, s.park\}@knu.ac.kr} \quad
{\small jw.jeong@keti.re.kr}
}

\begin{document}
\twocolumn[{%
\maketitle

\includegraphics[width=\linewidth]{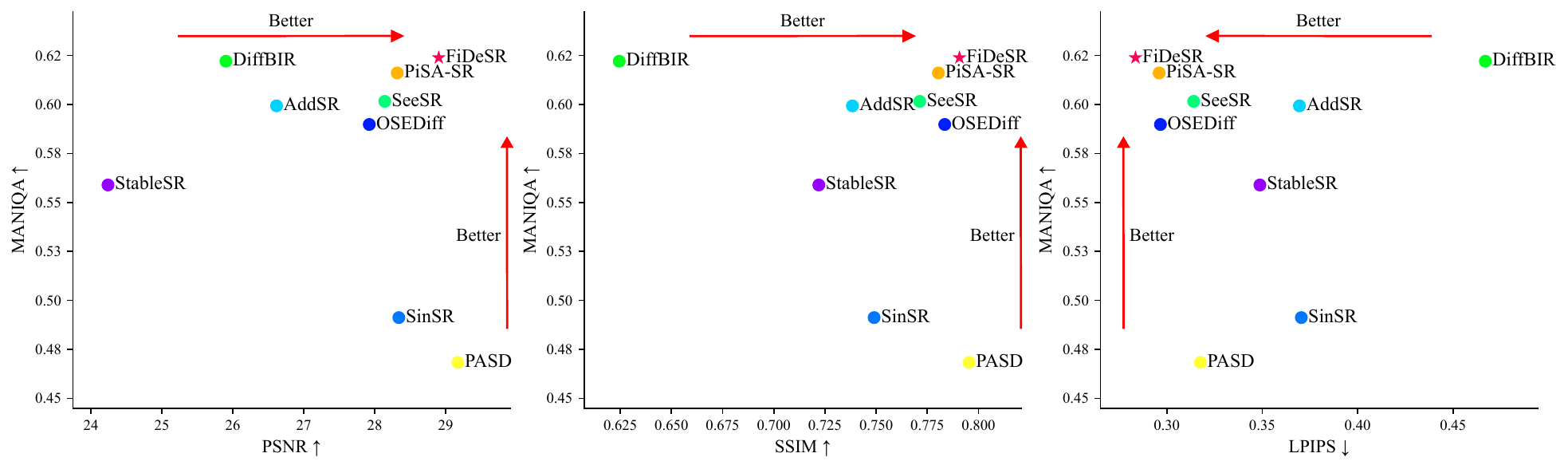}
\captionof{figure}{Performance comparison among Real-ISR methods on three perceptual–fidelity metric pairs: PSNR vs. MANIQA (left), SSIM vs. MANIQA (middle), and LPIPS vs. MANIQA (right). Higher MANIQA, PSNR, and SSIM values and lower LPIPS values indicate better performance. FiDeSR achieves superior perceptual quality while maintaining competitive fidelity across all three metric pairs. All methods are evaluated on the DRealSR dataset.}
\label{fig:teaser}
\vspace{1em}
}]
\let\thefootnote\relax\footnotetext{$^*$Equal contribution. $^\dagger$Corresponding author.}

\input{sec/0_abstract}    
\input{sec/1_intro}

\input{sec/2_related}

\input{sec/3_method}

\input{sec/4_experiments}
\input{sec/5_conclusion}

{
    \small
    \bibliographystyle{ieeenat_fullname}
    \bibliography{main}
}
\input{sec/X_suppl}


\end{document}

%% file: sec/0_abstract.tex
\begin{abstract}
Diffusion-based approaches have recently driven remarkable progress in real-world image super-resolution (SR). However, existing methods still struggle to simultaneously preserve fine details and ensure high-fidelity reconstruction, often resulting in suboptimal visual quality. In this paper, we propose FiDeSR, a high-fidelity and detail-preserving one-step diffusion super-resolution framework. During training, we introduce a detail-aware weighting strategy that adaptively emphasizes regions where the model exhibits higher prediction errors. During inference, low- and high-frequency adaptive enhancers further refine the reconstruction without requiring model retraining, enabling flexible enhancement control. To further improve the reconstruction accuracy, FiDeSR incorporates a residual-in-residual noise refinement, which corrects prediction errors in the diffusion noise and enhances fine detail recovery. FiDeSR achieves superior real-world SR performance compared to existing diffusion-based methods, producing outputs with both high perceptual quality and faithful content restoration. The source code will be released at: \url{https://github.com/Ar0Kim/FiDeSR}.
\end{abstract}

%% file: sec/1_intro.tex
\section{Introduction}
\label{sec:intro}

Image super-resolution (ISR) is the task of restoring a high-quality (HQ) image from its low-quality (LQ) input. Traditional ISR methods aim to restore downsampled LQ images  using simple bicubic interpolation \cite{yang2010image}. However, images captured in real-world environments suffer from unknown and complex degradations. To address these degradations, real-world image super-resolution (Real-ISR) has been developed to restore natural and realistic HQ images.

\begin{figure*}
    \centering
    \includegraphics[width=1\linewidth]{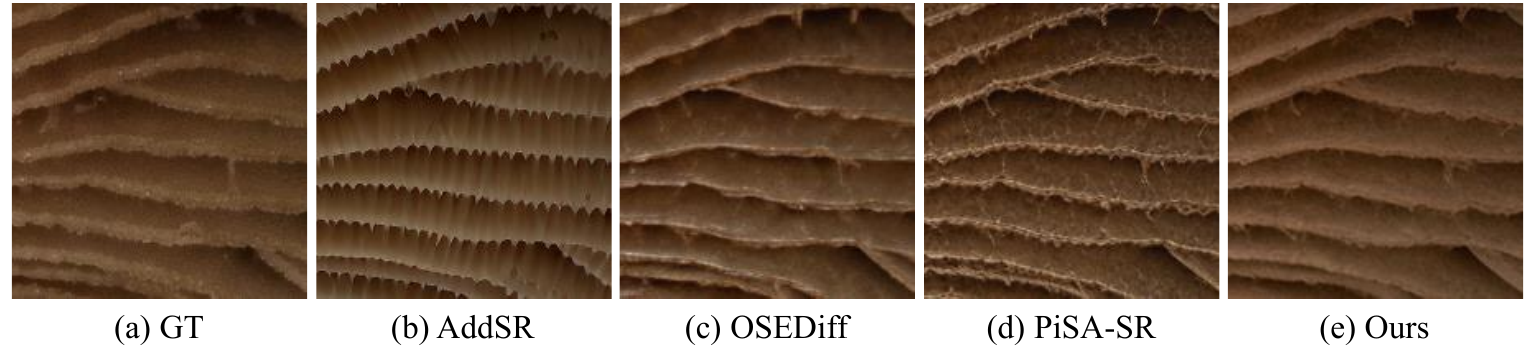}
    \caption{Example failure cases of diffusion-based Real-ISR methods. (b) AddSR introduces structural distortion and low-frequency inconsistency. (c) OSEDiff loses high-frequency details, producing over-smoothed texture. (d) PiSA-SR generates excessive details. In contrast, (e) our method achieves both high fidelity and detail-preserving.}
    \label{fig:detail}
\end{figure*}

Generative models have been utilized as a solution for producing realistic results in Real-ISR. Generative adversarial networks (GANs) \cite{goodfellow2014generative, ledig2017photo, wang2018esrgan} have shown success in generating realistic results. Recently, Diffusion Models (DMs) \cite{ho2020denoising, song2020score} have emerged as a solution for Real-ISR \cite{saharia2022image}, as their powerful image generation capability \cite{dhariwal2021diffusion} enables the restoration of realistic images under various degradation conditions. However, diffusion-based Real-ISR methods require iterative sampling steps, leading to high computational costs and long inference times \cite{yue2023resshift}. To address this issue, efficient one-step diffusion super-resolution models \cite{wang2024sinsr, wu2024one} have been proposed, which distill the multi-step diffusion process into a one-step sampling framework while maintaining reconstruction quality \cite{yin2024one}.

Among one-step diffusion-based Real-ISR methods, there remain two challenging problems: \textbf{(1) High-fidelity, (2) Detail restoration}. These approaches often suffer from structural distortion and low-frequency (LF) inconsistency due to VAE-based conditioning, making it difficult to preserve faithful image content while recovering fine high-frequency details, as observed in Fig.~\ref{fig:detail}(b) \cite{sun2023improving, arora2025guidesr}. 
In addition, one-step diffusion models also struggle with high-frequency (HF) detail loss.
Diffusion models generate HF details to compensate for the loss of HF information, which is degraded during the noise injection process of diffusion. In particular, latent diffusion models (LDMs) further amplify this loss, as the variational autoencoder (VAE) encoder compresses pixel-level representations into a compact latent space for efficient computation \cite{rombach2022high}.
While multi-step diffusion \cite{wang2024exploiting} enables sufficient or excessive detail restoration by generating HF details through iterative denoising, one-step diffusion reduces this process to a single step, resulting in insufficient HF detail restoration \cite{wu2024one, dong2025tsd}, as illustrated in Fig.~\ref{fig:detail}(c). 
Additionally, recent one-step diffusion models \cite{sun2025pixel, dong2025tsd} aim to predict the residual between the LQ and HQ latent representations. This residual learning strategy simplifies the SR process by allowing the model to focus on recovering HF information within the latent, leading to faster convergence and more efficient training. However, since the U-Net predicts only a single global residual at each stage, this approach results in unstable HF reconstruction and residual artifacts, as shown in Fig.~\ref{fig:detail}(d).

To overcome these limitations, we propose \textbf{a high-fidelity and detail-preserving one-step diffusion-based super-resolution (FiDeSR)} framework, which introduces three key components.
First, we employ a \textbf{Detail-aware Weighting (DAW)} strategy that utilizes frequency-based information to emphasize important detail-rich regions during training. DAW adaptively emphasizes regions where the current diffusion-based one-step model underperforms, rather than allowing it to overfit to already well-reconstructed areas. Second, we incorporate a \textbf{Latent Residual Refinement Block (LRRB)} to compensate for incomplete HF reconstruction and artifacts arising from coarse residual prediction. LRRB enhances the representational capacity of the model, while maintaining the efficiency of one-step diffusion. 
Finally, to simultaneously enhance perceptual detail and maintain fidelity, we introduce a \textbf{Latent Frequency Injection Module (LFIM)} that selectively injects frequency components extracted from the denoised latent according to spatial and channel characteristics.

As shown in Fig.~\ref{fig:teaser}, FiDeSR achieves superior perceptual quality while maintaining competitive fidelity on real-world super-resolution benchmarks, thereby demonstrating its ability to reconstruct fine details while preserving high structural fidelity compared to state-of-the-art diffusion-based Real-ISR methods. Our main contributions are summarized as follows:
\begin{itemize}
    \item We propose FiDeSR, a high-fidelity and detail-preserving one-step diffusion-based Real-ISR framework that effectively addresses challenges of structural fidelity degradation and insufficient high-frequency restoration in one-step diffusion models.
    \item We introduce three key technical components to form the FiDeSR framework: the Detail-aware Weighting (DAW) strategy, the Latent Residual Refinement Block (LRRB), and the Latent Frequency Injection Module (LFIM). These components are specifically designed to collectively address the challenges of high-fidelity and detail restoration in one-step diffusion-based Super-Resolution.
    \item FiDeSR achieves superior detail reconstruction and structural consistency on real-world SR benchmarks, outperforming state-of-the-art one-step and even competitive multi-step diffusion Real-ISR methods.
\end{itemize}

%% file: sec/2_related.tex
\section{Related Works}
\label{sec:related}
\subsection{Multi-step Diffusion-based Super-Resolution}
Diffusion models have been widely adopted in Real-ISR based on their strong generative priors. Early works applied pixel-space diffusion models such as DDPM \cite{ho2020denoising} for iterative refinement \cite{saharia2022image, yue2023resshift, li2022srdiff, qu2024xpsr, shang2024resdiff}, but these approaches suffer from high computational cost. To improve efficiency, later methods operate in the latent space of pretrained T2I diffusion models like Stable Diffusion \cite{rombach2022high}, often combined with PEFT techniques such as LoRA \cite{houlsby2019parameter, hu2022lora}. Representative approaches include StableSR \cite{wang2024exploiting}, DiffBIR \cite{lin2024diffbir}, PASD \cite{yang2024pixel}, and SeeSR \cite{wu2024seesr}, which leverage semantic prompts, feature injection, or restoration priors to guide the diffusion process. However, these methods still rely on multi-step denoising, resulting in slow inference.

\subsection{One-step Diffusion-based Super-Resolution}
To accelerate inference, recent studies compress the iterative diffusion process into a single forward pass via distillation. SinSR \cite{wang2024sinsr} adopts deterministic sampling and consistency-preserving loss to emulate multi-step behavior. AddSR \cite{xie2024addsr} applies adversarial diffusion distillation \cite{sauer2024adversarial} to reduce sampling steps, while OSEDiff \cite{wu2024one} uses VSD \cite{wang2023prolificdreamer} with LoRA-based fine-tuning to restore images directly from LQ latents. PiSA-SR \cite{sun2025pixel} introduces Dual-LoRA for controllable pixel and semantic refinement, and TSD-SR \cite{dong2025tsd} improves VSD stability using Target Score Distillation within SD3 \cite{esser2024scaling}. While TSD-SR performs well in terms of quantitative metrics, it exhibits noticeable grid artifacts, a limitation inherited from its SD3-based architecture. Although highly efficient, these one-step approaches still struggle with insufficient high-frequency detail, misalignment between LF structures and generated HF textures, and limitations of predicting a single global residual \cite{sun2025pixel, dong2025tsd}.

\subsection{Frequency-Domain Approaches for Super-Resolution}
Frequency-domain information has long been leveraged in SR to recover high-frequency details lost in degraded inputs \cite{lai2017deep, guo2017deep, qin2021fcanet, jiang2021focal, chen2023swinfsr, dai2024freqformer}. Recent diffusion-based SR methods have incorporated frequency cues to enhance generative restoration. RnG \cite{wang2023reconstruct} separates LF reconstruction and HF residual generation but relies on computationally expensive multi-step residual diffusion with a patch-wise step controller, which increases complexity and limits scalability. DiWa \cite{moser2024waving} and Frequency-Domain Refinement \cite{wang2024frequency} perform diffusion directly in the wavelet or frequency domain for finer HF recovery. FreeU \cite{si2024freeu} modulates diffusion features to strengthen HF reconstruction without retraining. Multi-Scale Generation Guidance \cite{shi2025multi} combines GAN supervision with DWT-based HF losses but still depends on iterative sampling and adversarial training. GuideSR \cite{arora2025guidesr} improves structural fidelity through full-resolution guidance in a one-step framework, yet remains biased toward fidelity and less effective in perceptual realism. TFDSR \cite{li2025timestep} incorporates frequency information into multi-step diffusion models by adapting frequency enhancement across diffusion timesteps within the denoising process. Despite these advances, most prior methods emphasize either high- or low-frequency components in isolation, or require additional overhead when addressing both. Thus, there is a need for a one-step diffusion framework that can effectively account for both frequency components and restore fine details and overall fidelity in a balanced manner.

%% file: sec/3_method.tex
\begin{figure*}
    \centering
    \includegraphics[width=0.95\linewidth]{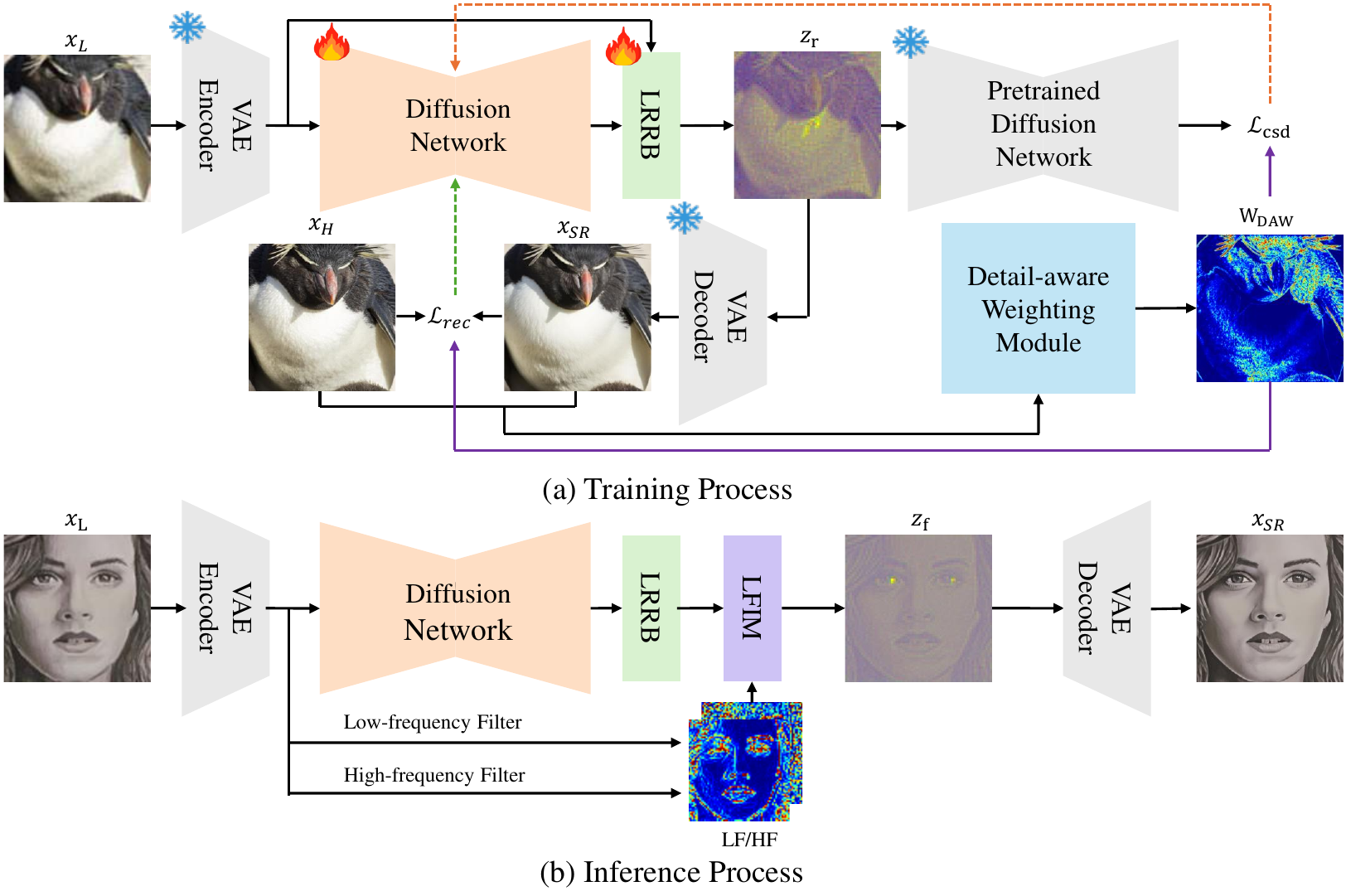}
    \caption{Overall framework of FiDeSR. (a) Training process: LQ image $x_L$ is encoded into a latent $z_L$, and the Diffusion Network predicts a coarse residual $r$, which is refined by LRRB to create refined latent $z_r$. The training loss is guided by the DAW module to emphasize fine structural details, and the refined latent ${z_r}$ is decoded by the VAE Decoder. (b) Inference process: Following the single-step diffusion process, the LQ latent $z_L$ is processed by the Diffusion Network to predict a residual, which is then refined by the LRRB. Refined $z_r$ is enhanced by frequency components through the LFIM, and finally decoded by the VAE Decoder to produce the Super-Resolution (SR) image $x_{SR}$. }
    \label{fig:framework}
\end{figure*}

\section{Method}

\subsection{Preliminaries}
The objective of Real-ISR is to restore a corresponding HQ image $x_H$ from a counterpart $x_L$ that has a complex and unknown degradation process.
The Real-ISR task is formulated as an optimization problem for a restoration network $G_\theta$ with parameters $\theta$, using $(x_L, x_H)$ pairs sampled from the training data distribution $D$. The objective is to find the optimal parameters $\theta^*$ that minimize the following composite function:
\begin{equation}
\theta^{*} = \arg\min_{\theta} \; \mathbb{E}_{(x_L, x_H) \sim D} 
\left[ \mathcal{L}_{rec}(x_{SR}, x_H) + \mathcal{L}_{reg}(x_{SR}) \right] .
\label{eq:1}
\end{equation}
This objective function attempts to balance two goals to achieve realistic restoration results. $\mathcal{L}_{rec}$ is the reconstruction loss, which enforces that the restored image $x_{SR}$ remains faithful to the HQ image $x_H$, typically measured using pixel-wise losses or perceptual metrics. This term is balanced by $\mathcal{L}_{reg}$, regularization loss, which guides the restored image $x_{SR}$ to follow the statistical properties of natural images. How to define and optimize the $\mathcal{L}_{reg}$ term in Eq. 1 has become the core challenge in Real-ISR.

Recently, Real-ISR methods leverage diffusion-based generative priors to satisfy this $\mathcal{L}_{reg}$ for more realistic restoration.
To reduce the high computational cost of multi-step diffusion inference, one-step diffusion SR methods compress the iterative denoising into a single step.
In particular, PiSA-SR \cite{sun2025pixel} performs restoration directly in the latent space of a pretrained VAE, where a U-Net predicts a global residual to bridge the latent representations between the LQ and HQ images in a single step. This residual-based restoration can be generally formulated as:
\begin{equation}
    z_0=z_L - r,
    \label{eq:2}
\end{equation}
where $z_L$ denotes the latent representation of the LQ image, $r$ is the residual predicted by the network, and $z_0$ denotes the restored latent obtained by applying the predicted residual to $z_L$.

\subsection{Overview of FiDeSR}
As shown in Fig.~\ref{fig:framework}(a), our FiDeSR framework aims to achieve high-fidelity, detail-preserving one-step super-resolution by integrating Detail-aware Weighting (DAW) and latent residual refinement block (LRRB) into the one-step diffusion process. Given a LQ input image $x_L$, we first encode it into the latent representation $z_L$ using a pretrained VAE encoder.
U-Net student network $G_\theta$ predicts a coarse residual $r$ to approximate the degradation between $z_L$ and its HQ counterpart $z_H$. This initial residual is further refined by our LRRB, which learns an adaptive correction $\Delta r$ based on both $z_L$ and $r$. The refined residual $r'$ is then used to reconstruct the refined latent $z_r$, which is decoded by the VAE decoder to produce the restored image.
During training, FiDeSR employs a DAW module to adaptively weight losses based on a spatial detail map computed from Sobel, Laplacian, and Variance filters. By emphasizing regions rich in edges and textures, DAW guides the model to focus on fine structural recovery and visually important details, leading to sharper and more faithful restoration.

\subsection{Detail-aware Weighting}
To adaptively emphasize regions that contain rich structural details and perceptually important textures, we introduce the DAW module. Unlike conventional frequency-based methods that explicitly decompose images in the Fourier domain, our DAW dynamically adjusts the loss by directly extracting HF details in the spatial domain. We construct a detail map $D$ to guide this weighting process. Specifically, multiple spatial operators including Sobel, Laplacian, and local variance filters are applied to the HQ image $x_H$ to capture edge sharpness, local contrast, and texture variance. The detail map is defined as the mean response of these operators:
\begin{equation}
    D = \frac{Sobel(x_{H}) + Laplacian(x_H) + Variance(x_H)}{3}.
\end{equation}

In addition, we compute an error map $E$ between the restored image $x_{SR}$ and the ground-truth image $x_H$, which captures both pixel-level and perceptual discrepancies. The pixel-wise error $E_{{pix}}$ is computed using the L1 difference, and the perceptual error $E_{{perc}}$ is computed using LPIPS. The pixel-wise error map $E_{{pix}}$ is defined as:
\begin{equation}
E_{{pix}} = |x_{SR} - x_H|.
\end{equation}
The perceptual error map $E_{{perc}}$ is obtained using an LPIPS network:
\begin{equation}
E_{{perc}} = {LPIPS}(x_{SR}, x_H).
\end{equation}
The final error map is computed as a weighted combination of pixel-level and perceptual errors:
\begin{equation}
E = (1 - p)\, E_{{pix}} + p \, E_{{perc}},
\end{equation}
where $p$ controls the contribution of perceptual discrepancy.

The DAW module integrates these two components to produce a difficulty weight map $W_{DAW}$, which represents the spatial difficulty of each pixel. This map is computed by element-wise multiplying the detail map $D$ and the error map $E$:
\begin{equation}
    W_{DAW} = D \odot E.
\end{equation}

This map adaptively scales the loss contribution of each spatial position, assigning higher weights to visually complex or detail-rich regions. We apply this weighting strategy to both the reconstruction loss and the classifier score distillation (CSD) loss \cite{sun2025pixel, wang2023prolificdreamer}, enabling the model to emphasize semantically and structurally difficult regions across all spatial objectives. By applying this strategy, our method effectively focuses the difficulty-aware weighting on spatially HQ regions, which guides the model to synthesize more accurate HF details, resulting in sharper textures, improved structural consistency, and enhanced visual realism.
\begin{figure}
    \centering
    \includegraphics[width=1\linewidth]{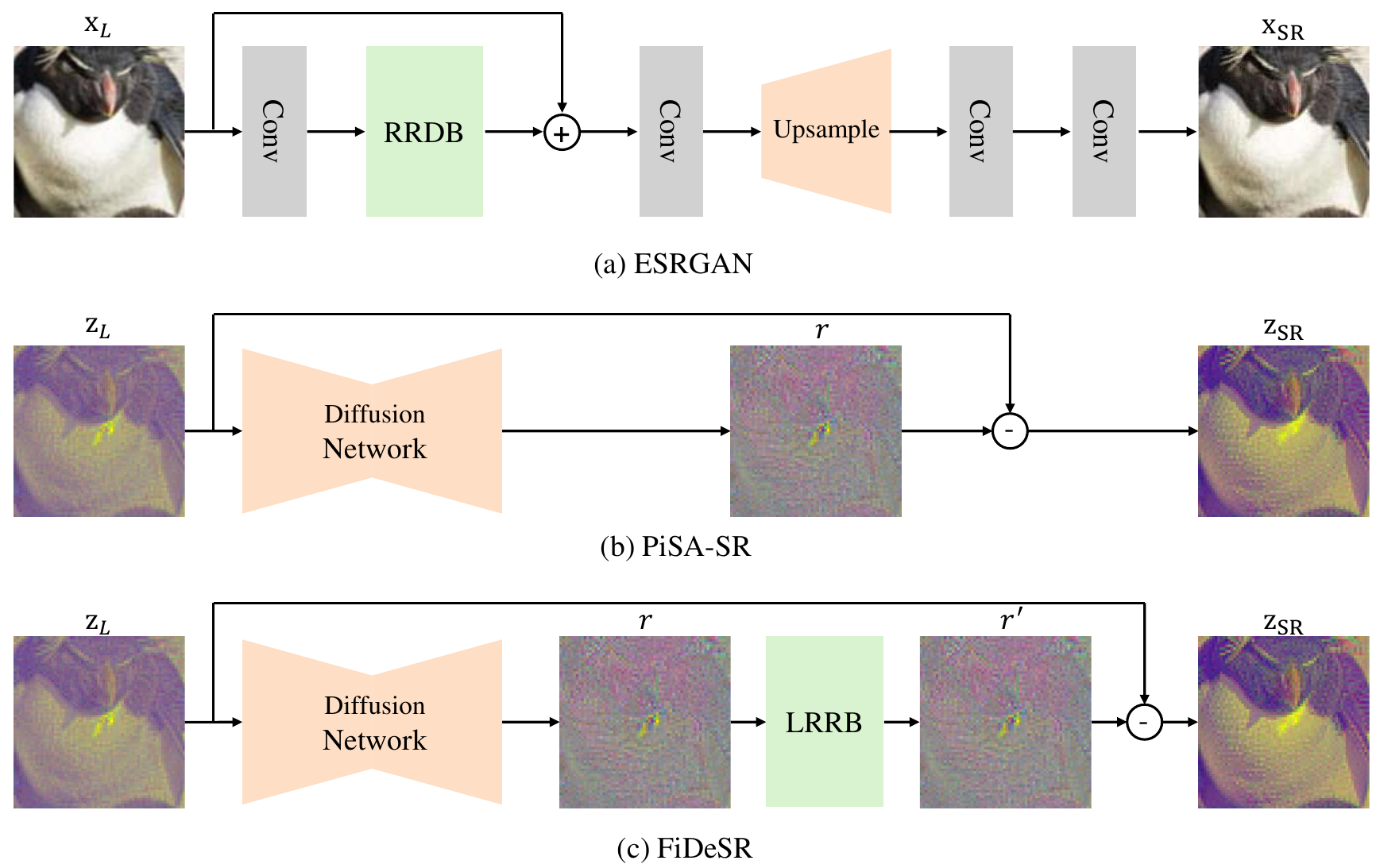}
    \caption{Comparison of the pipeline among ESRGAN, PiSA-SR, and our FiDeSR. (a) ESRGAN employs RRDB-based restoration in pixel space. (b) PiSA-SR uses a one-step diffusion network to predict a global residual in the latent space. (c) Our FiDeSR introduces latent residual refinement block (LRRB) that progressively refines the residual.}
    \label{fig:lrrb}
\end{figure}

\subsection{Latent Residual Refinement Block}
Existing one-step diffusion-based super-resolution methods (e.g., PiSA-SR \cite{sun2025pixel}) lack an iterative correction process, often resulting in unstable latent residual prediction and insufficient high-frequency recovery. 
To address this limitation, we introduce the LRRB, as shown in Fig.~\ref{fig:lrrb}(c).

LRRB operates in the latent space to refine the residual predicted by the diffusion U-Net before it is subtracted from the LQ latent $z_L$. 
Built upon the Residual-in-Residual Dense Block (RRDB) \cite{wang2018esrgan}, LRRB differs from pixel-domain residual refinement (e.g., ESRGAN), which assumes stable residual prediction. Instead, it explicitly targets diffusion-induced instability in latent residual prediction.

The LRRB takes the concatenation of the LQ latent $z_L$ and the initial residual of the U-Net $r$ as input. This combined latent is first mapped to an intermediate latent space by an initial 1x1 convolution, then passes through several dense block modules. Finally, a 1x1 convolution remaps these latents to the original residual channel dimension, generating the predicted correction value $\Delta r$.

This predicted correction $\Delta r$ is added to the original residual of the U-Net $r$ to create a refined residual $r'$:

\begin{equation}
r' = r + \Delta r .
\label{eq:3}
\end{equation}
This refined residual $r'$ is ultimately subtracted from the LQ latent $z_L$ to compute the refined latent $z_r$:

\begin{equation}
z_r = z_L - r' .
\label{eq:4}
\end{equation}
This refined latent $z_r$ is then passed to the VAE decoder to generate the final SR image.

This LRRB effectively advances the conventional simple residual subtraction method, as shown in Eq.~\ref{eq:2}, into a more powerful learning-based refinement stage. Instead of relying on U-Net initial prediction $r$, this output is treated as a strong initial estimate. LRRB then learns to predict an optimal correction $\Delta r$ based on both $r$ and the context of $z_L$. This two-step process, shown in Eq.~\ref{eq:3} and ~\ref{eq:4}, provides the model with greater flexibility to make more precise and complex adjustments to the residual. This learned refinement ultimately leads to a more accurate refined latent $z_r$ and a corresponding improvement in reconstruction quality.

\subsection{Training Losses}
The FiDeSR framework is optimized with a composite objective that enforces both pixel-level fidelity and semantic consistency. The total training loss consists of two key components: the reconstruction loss $\mathcal{L}_{rec}$ and the regularization loss $\mathcal{L}_{reg}$. Each of these loss terms is spatially modulated by the difficulty weight map $W_{DAW}$. This weighting guides the network to focus more on edges, textures, and fine details, resulting in sharper and more realistic reconstructions. 
The reconstruction loss $\mathcal{L}_{rec}$ is formulated as a sum of the pixel-wise MSE loss and the perceptual LPIPS loss, both spatially weighted by our $W_{DAW}$ map. To account for the different resolutions, the MSE loss is weighted by $W_{DAW}$ at the original image resolution, while the LPIPS loss is weighted by $W'_{DAW}$, which is an interpolated version of $W_{DAW}$ to match the LPIPS feature map dimensions:
\begin{equation}
\begin{split}
\mathcal{L}_{rec} = 
&\; \lambda_{mse} \, \mathbb{E}\!\left[ W_{DAW} \cdot (x_{SR} - x_H)^2 \right] \\
&+ \lambda_{lpips} \, \mathbb{E}\!\left[ W'_{DAW} \cdot \mathcal{L}_{LPIPS\_map} \right].
\end{split}
\label{eq:5}
\end{equation}

The regularization term $\mathcal{L}_{reg}$ is implemented as a CSD loss, which distills the semantic priors from a pretrained diffusion model to guide restored images towards natural and semantically consistent distributions with improved stability and efficiency. We modified the $\mathcal{L}_{CSD}$ function to accept the $W_{DAW}$ map as an argument. This allows the module to internally apply its semantic guidance with a focus on the critical regions identified by the map:
\begin{equation}
\mathcal{L}_{reg} = \lambda_{reg} \cdot \mathcal{L}_{CSD}(z_r, {prompt}, W_{DAW}) .
\label{eq:6}
\end{equation}

The total training objective $\mathcal{L}_{total}$ combines these components, following the structure established in Eq. ~\ref{eq:1}:
\begin{equation}
\mathcal{L}_{total} = \mathcal{L}_{rec} + \mathcal{L}_{reg} .
\label{eq:7}
\end{equation}
The model is guided by $W_{DAW}$ to synthesize more accurate HF details, resulting in sharper textures, improved structural consistency, and enhanced visual realism.

\subsection{LF/HF Adaptive Enhancers for Inference}
As illustrated in Fig.~\ref{fig:framework}(b), the inference process of FiDeSR restores a LQ input $x_L$ into a SR output $x_{SR}$ by combining the diffusion process with frequency-based detail enhancements. 
Following the diffusion process, the latent feature $z_{{L}}$ encoded by the VAE Encoder performs single-step restoration through the U-Net. The initial residual ($\mathbf{r}$) predicted by the U-Net is refined through the LRRB, resulting in a refined residual $r'$. This refined residual is then subtracted from $z_{{L}}$ to compute the refined latent $z_r$.

The refined latent $z_r$ is decomposed into low-frequency $\mathbf{\Delta}_{{LP}}$ and high-frequency $\mathbf{\Delta}_{{HP}}$ components by FFT-based Butterworth filters.
These frequency components are selectively injected into $z_r$ through the latent frequency injection module (LFIM).
The LFIM consists of a spatial gate $M_{sp}$, which identifies detailed and flat regions based on a detail map (Sobel, Laplacian, Variance) derived from the LQ image $x_L$, and a channel gate $M_{ch}$, which analyzes the frequency energy ratio of each latent channel. This selective injection strategy focuses on low-frequency enhancement in structure and high-frequency enhancement in texture, maximizing detail recovery. The final enhanced latent $z_{f}$ is passed to the VAE Decoder to generate the final HQ image $x_{SR}$ in pixel space.

%% file: sec/4_experiments.tex
\begin{table*}[t]
\centering
\caption{Quantitative comparison with state-of-the-art DM-based SR methods on synthetic and real-world test datasets. The number of diffusion inference steps is indicated by 's'. The best and second best results of each metric are highlighted in {\color{red}\textbf{red}} and {\color{blue}\textbf{blue}}, respectively. The third best results are indicated by an \textbf{\underline{underline}}.}
\label{tab:sr_results}

\scriptsize 
\setlength{\tabcolsep}{4pt} 
\renewcommand{\arraystretch}{0.90} 

\resizebox{\textwidth}{!}{%
\begin{tabular}{llccccccccc}
\toprule
\textbf{Dataset} & \textbf{Method} & \textbf{PSNR $\uparrow$} & \textbf{SSIM $\uparrow$} & \textbf{LPIPS $\downarrow$} & \textbf{DISTS $\downarrow$} & \textbf{CLIPIQA $\uparrow$} & \textbf{NIQE $\downarrow$} & \textbf{MUSIQ $\uparrow$} & \textbf{MANIQA $\uparrow$} & \textbf{FID $\downarrow$} \\
\midrule
\multirow{9}{*}{DRealSR} & StableSR-200s & 27.93 & 0.7491 & 0.3306 & 0.2290 & 0.6197 & 6.3865 & 58.5502 & 0.5569 & 147.48 \\
 & SeeSR-50s & 28.14 & 0.7713 & 0.3141 & 0.2298 & 0.6889 & 6.4633 & 64.7302 & 0.6016 & 146.98 \\
 & DiffBIR-50s & 25.90 & 0.6245 & 0.4669 & 0.2298 & 0.6333 & \textbf{\underline{6.3275}} & \color{red}\textbf{66.1334} & \color{blue}\textbf{0.6221} & 180.34 \\
 & PASD-20s & \color{red}\textbf{29.17} & \color{red}\textbf{0.7954} & 0.3176 & 0.2220 & \color{blue}\textbf{0.7071} & 7.7138 & 50.4422 & 0.4683 & 143.08 \\
 & AddSR-4s & 26.62 & 0.7384 & 0.3695 & 0.2651 & \color{red}\textbf{0.7121} & 7.8536 & 65.0625 & 0.5993 & 167.69 \\
 & OSEDiff-1s & 27.92 & \textbf{\underline{0.7835}} & \textbf{\underline{0.2967}} & \color{blue}\textbf{0.2162} & 0.6956 & 6.4389 & 64.6979 & 0.5898 & \textbf{\underline{135.45}} \\
 & SinSR-1s & \textbf{\underline{28.34}} & 0.7490 & 0.3705 & 0.2487 & 0.6445 & 7.0300 & 55.3668 & 0.4912 & 175.96 \\
 & PiSA-SR-1s & 28.32 & 0.7804 & \color{blue}\textbf{0.2960} & \textbf{\underline{0.2169}} & 0.6918 & \color{red}\textbf{6.1803} & \color{blue}\textbf{66.1112} & \textbf{\underline{0.6161}} & \color{blue}\textbf{130.48} \\
 & \textbf{FiDeSR-1s} & \color{blue}\textbf{28.90} & \color{blue}\textbf{0.7907} & \color{red}\textbf{0.2836} & \color{red}\textbf{0.2112} & \textbf{\underline{0.6974}} & \color{blue}\textbf{6.2014} & \textbf{\underline{65.7820}} & \color{red}\textbf{0.6239} & \color{red}\textbf{127.97} \\
\midrule
\multirow{9}{*}{RealSR} & StableSR-200s & 24.73 & 0.7084 & 0.3050 & 0.2156 & 0.6379 & \textbf{\underline{5.6251}} & 65.4419 & 0.6259 & 130.43 \\
 & SeeSR-50s & 25.21 & 0.7216 & 0.3004 & 0.2218 & 0.6674 & 5.5929 & 69.6929 & 0.6435 & 125.09 \\
 & DiffBIR-50s & 24.83 & 0.6501 & 0.3650 & 0.2399 & \color{blue}\textbf{0.7053} & 5.8404 & 69.2781 & 0.6502 & 130.52 \\
 & PASD-20s & \color{red}\textbf{26.63} & \color{red}\textbf{0.7660} & \textbf{\underline{0.2869}} & \color{blue}\textbf{0.2032} & 0.4857 & 5.9815 & 60.0287 & 0.5390 & 124.52 \\
 & AddSR-4s & 22.54 & 0.6419 & 0.3867 & 0.2718 & \color{red}\textbf{0.7298} & 6.5640 & \color{red}\textbf{71.4404} & \color{red}\textbf{0.6751} & 156.68 \\
 & OSEDiff-1s & 25.15 & 0.7341 & 0.3194 & 0.2127 & 0.6686 & 5.6364 & 69.0810 & 0.6335 & \color{blue}\textbf{123.49} \\
 & SinSR-1s & \color{blue}\textbf{26.29} & 0.7350 & 0.3194 & 0.2344 & 0.6181 & 6.2854 & 60.7494 & 0.5420 & 134.27 \\
 & PiSA-SR-1s & 25.50 & \textbf{\underline{0.7418}} & \color{blue}\textbf{0.2672} & \textbf{\underline{0.2044}} & 0.6697 & \color{blue}\textbf{5.5054} & \color{blue}\textbf{70.1462} & \textbf{\underline{0.6551}} & \textbf{\underline{124.18}} \\
 & \textbf{FiDeSR-1s} & \textbf{\underline{26.02}} & \color{blue}\textbf{0.7457} & \color{red}\textbf{0.2626} & \color{red}\textbf{0.1965} & \textbf{\underline{0.6896}} & \color{red}\textbf{5.3194} & \textbf{\underline{69.8245}} & \color{blue}\textbf{0.6681} & \color{red}\textbf{109.68} \\
 \midrule
\multirow{9}{*}{DIV2K} & StableSR-200s & 23.27 & 0.5733 & 0.3106 & 0.2045 & 0.6773 & 4.7621 & 65.8259 & 0.6178 &\color{blue}\textbf{24.43}\\
 & SeeSR-50s & 23.73 & 0.6057 & 0.3193 & \textbf{\underline{0.1966}} & 0.6864 & 4.7949 & 68.3976 & 0.6200 & 25.80 \\
 & DiffBIR-50s & 23.14 & 0.5441 & 0.3669 & 0.2209 & \color{blue}\textbf{0.7299} & 4.9920 & \color{red}\textbf{69.8698} & \color{red}\textbf{0.6440} & 32.71 \\
 & PASD-20s & \color{red}\textbf{24.41} & \color{red}\textbf{0.6252} & 0.3794 & 0.2218 & 0.5567 & 5.4273 & 61.2676 & 0.5362 & 31.53 \\
 & AddSR-4s & 22.38 & 0.5554 & 0.3815 & 0.2340 & \color{red}\textbf{0.7538} & 5.8168 & \textbf{\underline{69.1623}} & 0.6300 & 35.07 \\
 & OSEDiff-1s & 23.72 & \textbf{\underline{0.6109}} & \textbf{\underline{0.2942}} & 0.1975 & 0.6681 & \textbf{\underline{4.7104}} & 67.9670 & 0.6131 & 26.33 \\
 & SinSR-1s & \color{blue}\textbf{24.40} & 0.6018 & 0.3244 & 0.2067 & 0.6492 & 6.0002 & 62.8673 & 0.5393 & 35.59 \\
 & PiSA-SR-1s & 23.87 & 0.6058 & \color{blue}\textbf{0.2823} & \color{blue}\textbf{0.1934} & \textbf{\underline{0.6928}} & \color{red}\textbf{4.5563} & \color{blue}\textbf{69.6799} & \textbf{\underline{0.6375}} & \textbf{\underline{25.09}} \\
 & \textbf{FiDeSR-1s} & \textbf{\underline{24.33}} & \color{blue}\textbf{0.6250} & \color{red}\textbf{0.2678} & \color{red}\textbf{0.1845} & 0.6873 & \color{blue}\textbf{4.6644} & 68.8672 & \color{blue}\textbf{0.6384} & \color{red}\textbf{23.30} \\
\bottomrule
\end{tabular}%
}
\end{table*}

\begin{figure*}
    \centering
    \includegraphics[width=0.8\linewidth]{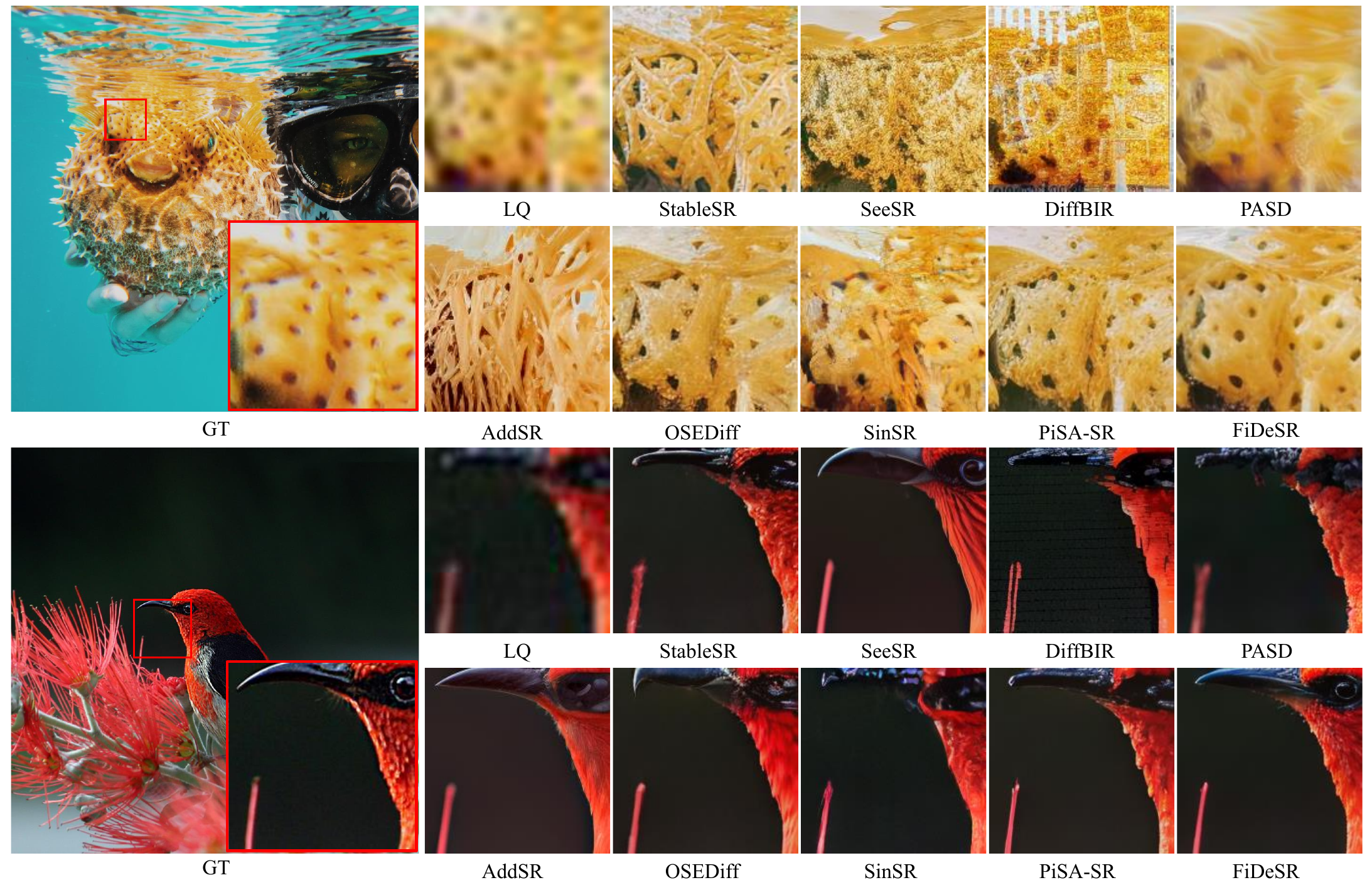}
    \caption{Qualitative comparisons with state-of-the-art DM-based SR methods.}
    \label{fig:qualitative}
\end{figure*}

\section{Experiments}
\subsection{Experimental Settings}
\noindent\textbf{Training Datasets.}
We employed LSDIR \cite{li2023lsdir}, DIV2K \cite{agustsson2017ntire}, Flickr2K \cite{timofte2017ntire}, and the first 10K images from FFHQ \cite{karras2019style} as training data. We generated LQ-HQ pairs for Real-ISR training by degrading HR images using the Real-ESRGAN \cite{wang2021real} degradation pipeline.

\noindent\textbf{Test Datasets.}
We evaluate our method on both synthetic and real-world datasets. The synthetic test dataset consists of 3,000 images of 512x512, cropped from DIV2K-validation \cite{agustsson2017ntire} and degraded into 128x128 LQ images using the Real-ESRGAN degradation pipeline. The real-world datasets are obtained by center-cropping images from RealSR \cite{cai2019toward} and DRealSR \cite{wei2020component}, where the LQ and HQ images have resolutions of 128x128 and 512x512.

\noindent\textbf{Evaluation Metrics.}
To comprehensively evaluate the performance of our method, we employ both full-reference and no-reference metrics. For full-reference metrics, PSNR and SSIM \cite{wang2004image} are used to measure pixel-level fidelity between restored and GT images. LPIPS \cite{zhang2018unreasonable} and DISTS \cite{ding2020image} are adopted to evaluate perceptual quality by comparing deep feature similarities. FID \cite{heusel2017gans} is utilized to evaluate the distributional distance between the restored and GT image sets. No-reference metrics are used to evaluate perceptual quality without relying on GT images, including CLIPIQA \cite{wang2023exploring}, NIQE \cite{zhang2015feature}, MUSIQ \cite{ke2021musiq}, and MANIQA \cite{yang2022maniqa}.

\noindent\textbf{Implementation details.}
Our method is implemented based on the Stable Diffusion 2.1-base \cite{StabilityAI_SD}. The pretrained VAE and U-Net are kept frozen, and we add trainable LoRA layers to the U-Net for fine-tuning, setting the LoRA rank to 8.
We train our model on 2 NVIDIA H100 GPUs with a batch size of 8 for 200K training steps. 
We used the AdamW optimizer \cite{loshchilov2017decoupled} with a learning rate of $5 \times 10^{-5}$. We utilize RAM \cite{zhang2024recognize} to extract text prompts. In all experiments, the loss weights are set to $\lambda_{\mathrm{mse}}=1$ and $\lambda_{\mathrm{lpips}}=2$.

\subsection{Comparison with existing methods}

\noindent\textbf{Quantitative comparisons.}
Table~\ref{tab:sr_results} provides a quantitative comparison with recent diffusion-based SR methods on synthetic and real-world benchmarks. Across all datasets, FiDeSR shows consistently reliable performance, maintaining a balanced improvement in both full-reference and no-reference metrics while requiring only a single diffusion step. In particular, FiDeSR achieves consistently better scores in perceptual similarity measures such as LPIPS and DISTS, indicating improved reconstruction and reduced artifacts compared to other one-step and multi-step diffusion models. At the same time, FiDeSR also demonstrates competitive no-reference performance on metrics such as CLIP-IQA, MUSIQ, and MANIQA, suggesting that the restored images better align with natural image statistics and exhibit more coherent visual structure. This simultaneous improvement on both reference-based and no-reference evaluation is challenging due to the inherent perception–distortion trade-off, yet FiDeSR manages to maintain a more stable balance than prior diffusion-based SR approaches. Furthermore, FiDeSR achieves consistently lower FID values than all existing methods including both one-step and multi-step methods, demonstrating the closest alignment with real-image distributions overall.  Overall, FiDeSR demonstrates consistently superior restoration performance across diverse real-world settings by effectively addressing the common limitations of one-step diffusion, overcoming insufficient high-frequency detail reconstruction and low-frequency structural inconsistency.

\noindent\textbf{Qualitative comparisons.}
Visual results are shown in Fig.~\ref{fig:qualitative}. Multi-step methods such as SeeSR generate rich details, but many of them appear unnatural, often introducing noise-like artifacts and compromising fidelity. AddSR tends to deviate from the ground-truth structure, producing distorted shapes and lacking fine local details. DiffBIR also exhibits noticeable noise artifacts, while PASD often produces blurry textures and degraded structural consistency. OSEDiff preserves some coarse structures but still suffers from reduced fidelity and insufficient high-frequency reconstruction. PiSA-SR improves perceptual sharpness but may introduce overly enhanced or inconsistent textures in complex regions. In contrast, FiDeSR restores both structural integrity and fine details more faithfully to the ground truth while simultaneously producing sharper textures and a more natural overall appearance, without introducing unwanted artifacts.

{\setlength{\textfloatsep}{6pt}
\setlength{\floatsep}{6pt}
\setlength{\intextsep}{6pt}
\setlength{\abovecaptionskip}{4pt}
\setlength{\belowcaptionskip}{0pt}

\subsection{Ablation study}
\begin{table}[t]
\centering
\caption{Ablation study on LRRB and DAW on the DIV2K dataset (evaluated before applying LFIM).}
\resizebox{0.48\textwidth}{!}{%
\begin{tabular}{lcccc}
\toprule
\textbf{Model} & \textbf{CLIPIQA} $\uparrow$ & \textbf{NIQE} $\downarrow$ & \textbf{MUSIQ} $\uparrow$ & \textbf{MANIQA} $\uparrow$ \\
\midrule
w/o LRRB, DAW & 0.6611 & 4.7381 & 67.6043 & 0.6237 \\
w/o LRRB      & 0.6641 & 4.7129 & 67.6280 & 0.6236 \\
w/o DAW       & 0.6626 & 4.7340 & 67.9515 & 0.6278 \\
\textbf{FiDeSR}      & \textbf{0.6699} & \textbf{4.6300} & \textbf{68.2869} & \textbf{0.6285} \\
\bottomrule
\end{tabular}
}
\label{tab:lf_hf_ablation}
\end{table}

\begin{table}[t]
\centering
\caption{High-frequency noise prediction error comparison between baseline and LRRB models. Lower MSE indicates better performance.}
\label{tab:real_sr_comparison}

\scriptsize
\setlength{\tabcolsep}{11pt} 
\renewcommand{\arraystretch}{0.9} 

\begin{tabular}{lcccc}
\toprule
\textbf{Method} & DIV2K & DRealSR & RealSR & Average \\
\midrule
baseline & 0.1051 & 0.1049 & 0.1029 & 0.1049 \\
LRRB & 0.1038 & 0.1028 & 0.1045 & 0.1032 \\
\midrule
Improvement & 1.24\% & 1.99\% & 1.62\% & 1.62\% \\
\bottomrule
\end{tabular}
\label{tab:errorreduction}
\end{table}

\noindent\textbf{Effectiveness of LRRB and DAW.} Table~\ref{tab:lf_hf_ablation} shows that both the LRRB and DAW modules contribute to perceptual quality. Removing either module consistently degrades the CLIPIQA, NIQE, MUSIQ, and MANIQA scores, indicating reduced detail preservation. When used together, FiDeSR achieves the best performance across all metrics, demonstrating the complementary effect of both modules in enhancing detail-aware restoration. The quantitative comparison in Table~\ref{tab:errorreduction} further  validates the benefit of incorporating the LRRB module. Across DIV2K, DRealSR, and RealSR datasets, the LRRB-equipped model consistently reduces the high-frequency noise prediction error compared to the baseline. More comprehensive results and analyses can be found in the supplementary materials.

\noindent\textbf{Ablation on LFIM (HF/LF).}
Table~\ref{tab:ablation_hflf_full} summarizes the effects of LFIM on both low- and high-frequency components. 
Applying LFIM on low-frequency features consistently improves PSNR and SSIM as the injection intensity increases, 
which confirms that reinforcing low-frequency information stabilizes global structure, tone, and illumination, thereby enhancing structural fidelity. 
In contrast, applying LFIM on high-frequency components substantially enhances perceptual quality metrics such as MUSIQ and MANIQA by selectively enhancing fine textures, edges, and high-frequency details. 
This demonstrates that the LFIM effectively complements structural refinement with perceptual sharpness, contributing to balanced and detail-aware super-resolution.

\begin{table}[t]
\centering
\caption{Ablation of LFIM (HF/LF) modules with different injection on the RealSR dataset.}

\scriptsize  

\setlength{\tabcolsep}{9pt}
\renewcommand{\arraystretch}{0.9}

\begin{tabular}{lcccc}
\toprule
\textbf{Method} & \textbf{PSNR} $\uparrow$ & \textbf{SSIM} $\uparrow$ & \textbf{MUSIQ} $\uparrow$ & \textbf{MANIQA} $\uparrow$ \\
\midrule
Baseline & 26.2542 & 0.7498 & 69.3610 & 0.6580 \\
\midrule
HF-0.1 & 26.1295 & 0.7476 & 69.6450 & 0.6639 \\
HF-0.2 & 25.9985 & 0.7452 & 69.8562 & 0.6692 \\
HF-0.3 & 25.8618 & 0.7428 & 70.0120 & 0.6737 \\
HF-0.4 & 25.7213 & 0.7403 & 70.1132 & 0.6774 \\
HF-0.5 & 25.5778 & 0.7378 & 70.1617 & 0.6803 \\
\midrule
LF-0.1 & 26.2905 & 0.7506 & 69.3177 & 0.6571 \\
LF-0.2 & 26.3066 & 0.7513 & 69.2337 & 0.6564 \\
LF-0.3 & 26.3221 & 0.7520 & 69.1431 & 0.6550 \\
LF-0.4 & 26.3377 & 0.7526 & 69.0521 & 0.6538 \\
LF-0.5 & 26.3448 & 0.7532 & 68.9473 & 0.6522 \\
\bottomrule
\end{tabular}

\label{tab:ablation_hflf_full}
\end{table}

}

%% file: sec/5_conclusion.tex
\section{Conclusion}
In this work, we presented FiDeSR, a high-fidelity and detail-preserving one-step diffusion super-resolution framework that addresses two challenges of existing one-step diffusion methods: structural fidelity degradation and insufficient high-frequency detail restoration. To overcome these challenges, FiDeSR introduces three components: detail-aware Weighting, latent residual refinement block, and the latent frequency injection module. Through extensive experiments, FiDeSR demonstrates state-of-the-art performance among one-step diffusion SR models and achieves consistent gains in both full-reference and no-reference metrics. Overall, FiDeSR highlights that one-step diffusion models can achieve high perceptual quality without sacrificing fidelity, when both frequency-aware guidance and residual refinement are properly integrated, which would open new directions for efficient real-world SR and extensions to video or multi-modal restoration tasks.
\section{Acknowledgment}
This work was supported in part by the National Research Foundation of Korea (NRF) grant funded by the Korean government (MSIT) (No. RS-2025-00520308) (50\%), in part by the Regional Innovation System \& Education (RISE) Glocal 30 program through the Daegu RISE Center, funded by the Ministry of Education (MOE) and Daegu, Republic of Korea (2025-RISE-03-001) (25\%), and in part by the BK21 FOUR project (AI-driven Convergence Software Education Research Program) funded by the Ministry of Education, School of Computer Science and Engineering, Kyungpook National University, Korea (41202420214871) (25\%).

%% file: sec/X_suppl.tex
\clearpage

\maketitlesupplementary

\appendix 

This supplementary material includes: comparison with user Study (Sec.~\ref{sec:user}), GAN-based Real-ISR (Sec.~\ref{sec:gan}), more visual comparisons (Sec.~\ref{sec:more}), complexity analysis (Sec. ~\ref{sec:complexity}), ablation studies (Sec.~\ref{sec: ablation}), implementation details of DAW (Sec.~\ref{sec:DAW_implement}) and LFIM (Sec.~\ref{sec:LFIM}), and comparison with Frequency-Aware Diffusion SR (Sec.~\ref{sec:freq}).

\section{User Study}
\label{sec:user}
\begin{figure*}
    \centering
    \includegraphics[width=0.9\linewidth]{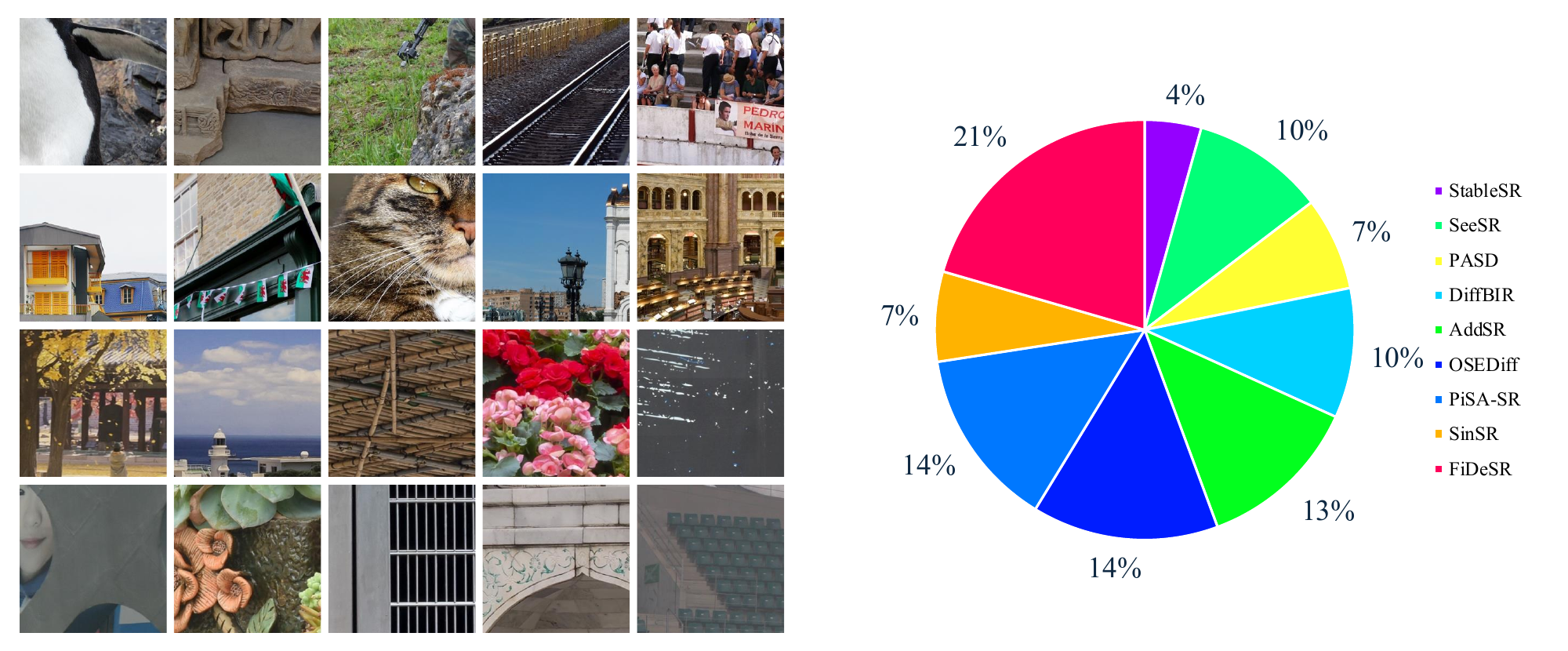}
    \caption{Ground-truth images used in the user study along with the voting results for their corresponding restored outputs.}
    \label{fig:placeholder}
\end{figure*}

As shown in Fig.~\ref{fig:placeholder}, we conduct a user study comparing our method with eight other diffusion-based SR approaches. A total of 20 images were randomly selected from the DIV2K \cite{agustsson2017ntire}, DRealSR \cite{wei2020component}, and RealSR \cite{cai2019toward} datasets, and 20 volunteers were asked to select the image that best balances perceptual realism and content fidelity. Although individual preferences varied, FiDeSR received the highest number of votes overall, demonstrating its superior perceptual quality compared with the competing methods.

\section{Comparison with GAN-based Real-ISR}
\label{sec:gan}
\begin{figure*}
    \centering
    \includegraphics[width=1\linewidth]{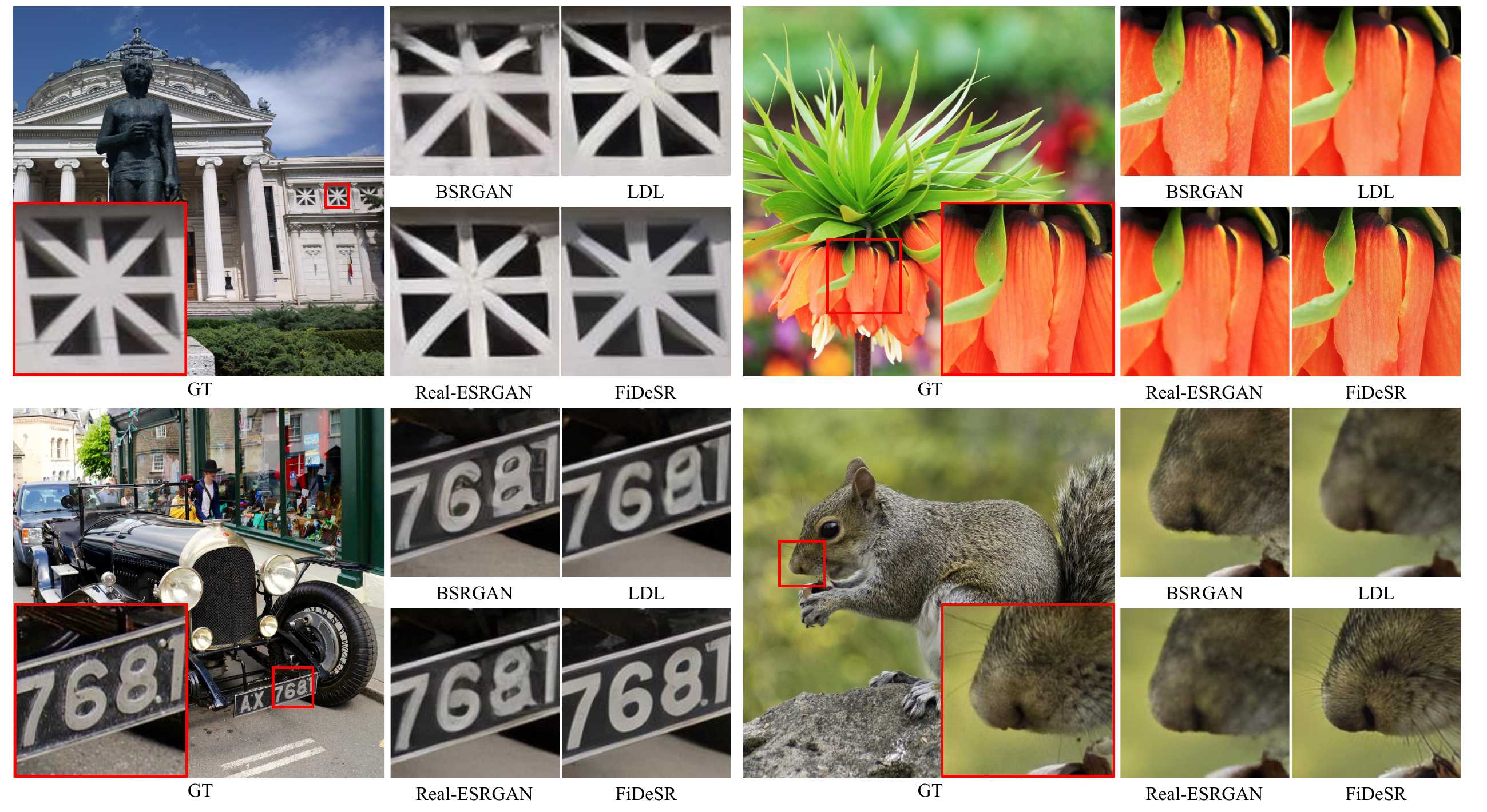}
    \caption{Qualitative comparisons between FiDeSR and GAN-based Real-ISR methods.}
    \label{fig:GAN}
\end{figure*}

We compare our method with GAN-based Real-ISR in Table~\ref{tab:sr_results_gan}. As shown in the table, our FiDeSR outperforms existing GAN-based methods \cite{wang2021real, liang2022details, zhang2021designing}. In particular, FiDeSR achieves superior results in LPIPS and DISTS, indicating improved perceptual similarity and reduced artifact formation compared to existing GAN-based methods.
FiDeSR shows strong results in no-reference metrics such as CLIP-IQA, NIQE, MUSIQ, and MANIQA, suggesting that the restored images look more realistic and maintain coherent visual structure.
In addition, we present visual comparisons to GAN-based Real-ISR methods in Fig.~\ref{fig:GAN}.
As shown in the figure, GAN-based models often suffer from over-smoothed textures, inaccurate details, and unstable structure reconstruction. FiDeSR produces more faithful textures and preserves geometric structures more consistently.

\section{More Visual Comparisons}
\label{sec:more}
\begin{figure*}
    \centering
    \includegraphics[width=0.9\linewidth]{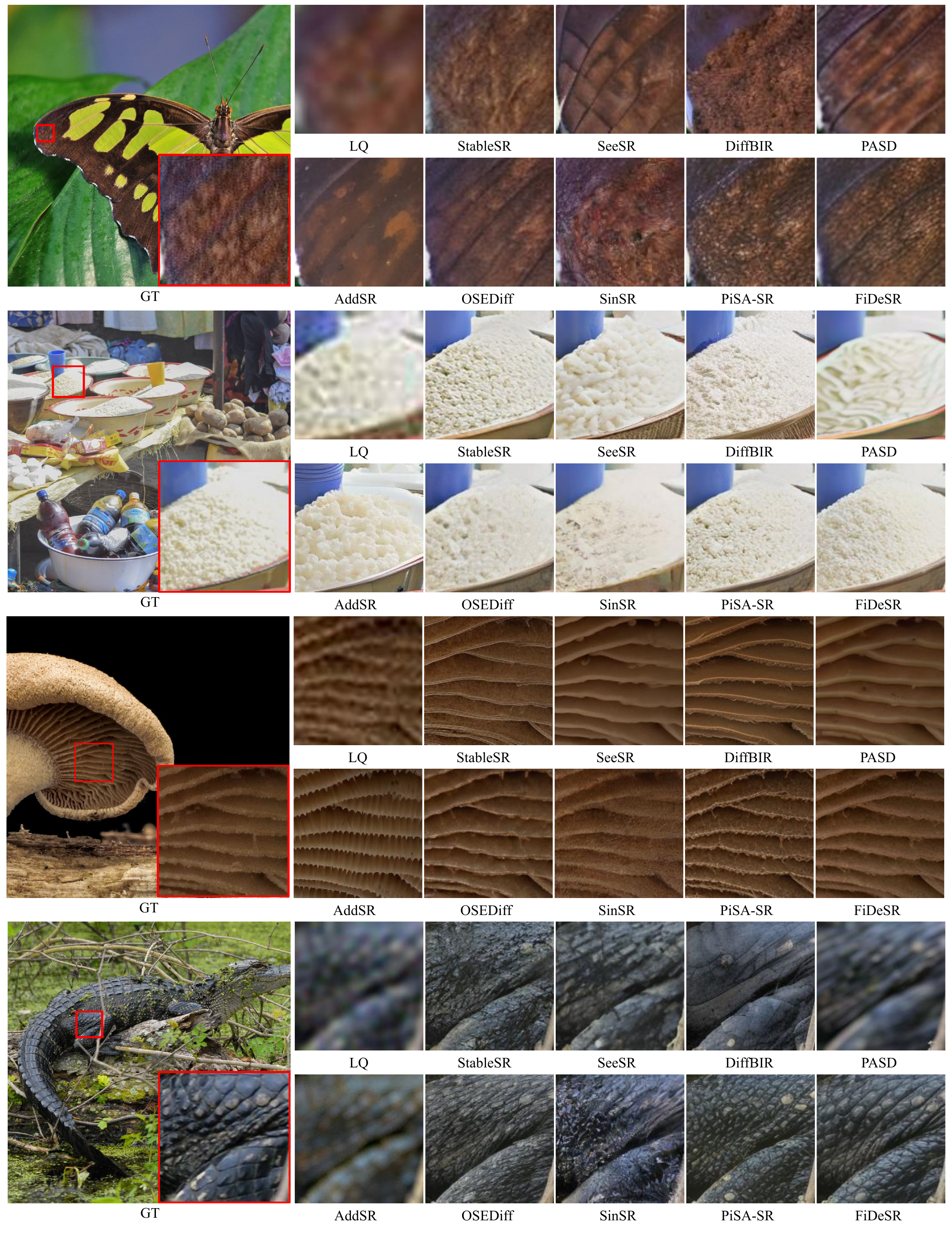}
    \caption{Qualitative comparisons between FiDeSR and different diffusion-based methods on DIV2K dataset. FiDeSR effectively reconstructs fine details while preserving overall image fidelity.}
    \label{fig:visual_DIV}
\end{figure*}

\begin{figure*}
    \centering
    \includegraphics[width=0.9\linewidth]{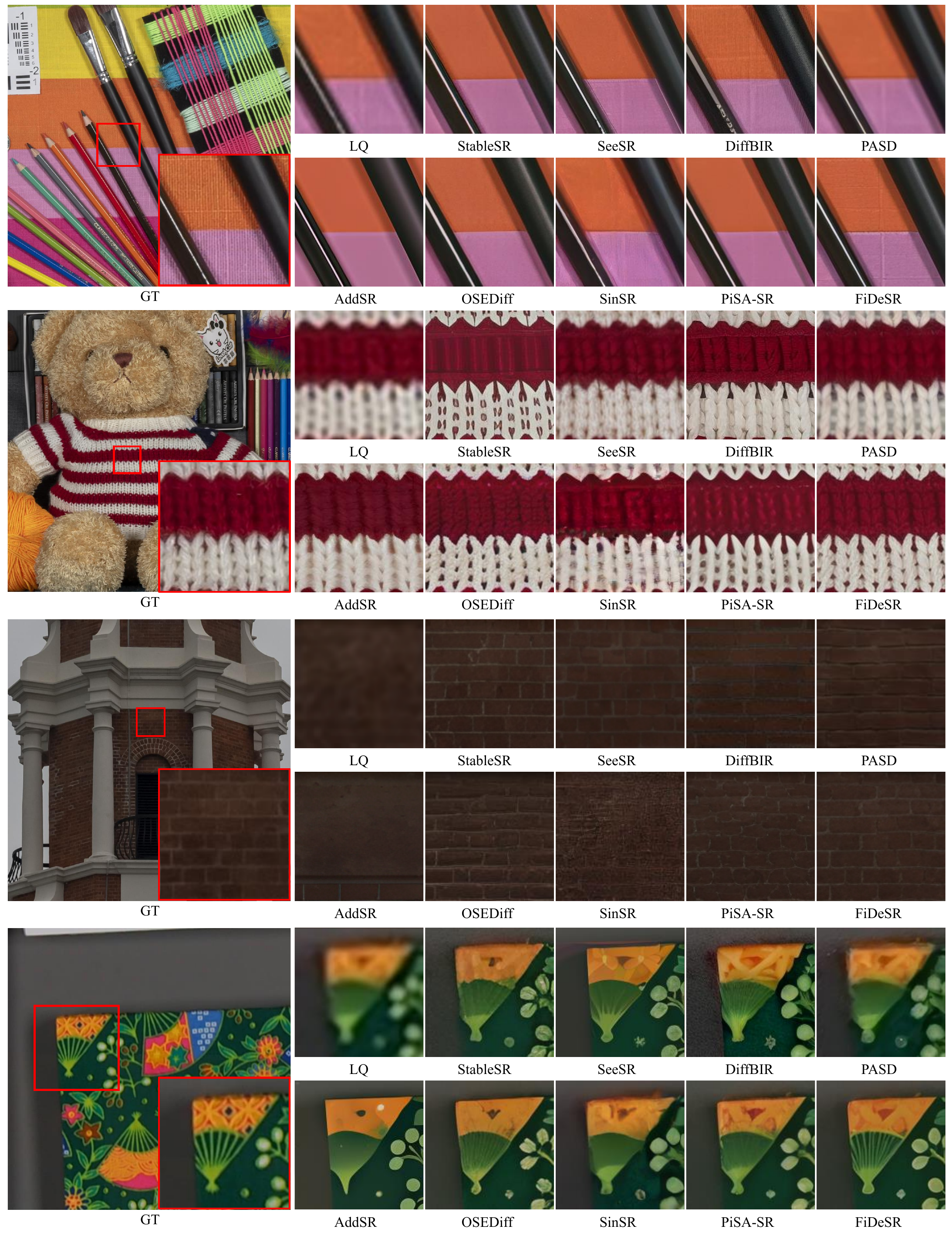}
    \caption{Qualitative comparisons between FiDeSR and different diffusion-based methods on DRealSR and RealSR dataset. FiDeSR effectively reconstructs fine details while preserving overall image fidelity.}
    \label{fig:visual_real}
\end{figure*}

As shown in Fig.~\ref{fig:visual_DIV} and Fig.~\ref{fig:visual_real}, we provide additional visual comparisons between FiDeSR and other diffusion-based SR models. FiDeSR consistently reconstructs fine textures and intricate patterns even from heavily degraded low-quality inputs, producing results that are both highly realistic and structurally faithful compared with other methods.

\section{Complexity Analysis}
\label{sec:complexity}
We compare the number of parameters and the inference time of competing diffusion model-based SR models in Table~\ref{tab:inference_step}. The inference time is measured on the $\times 4$ SR task using $128 \times 128$ LQ images and a single NVIDIA H100 80GB GPU. Although FiDeSR incorporates both the LRRB and LFIM modules and is designed to produce perceptually faithful and realistic reconstructions, it still achieves competitive inference speed compared with existing one-step SR models.
In addition, compared to FiDeSR without LRRB, introducing LRRB increases the parameter count by only 0.01B (0.8\% of 1.29B) and adds a runtime overhead of 0.0063 s (8.1\% of 0.078 s).

\section{Ablation studies}
\label{sec: ablation}

\subsection{LRRB High-Frequency Noise Prediction Refinement Visualization}

By employing the LRRB, we demonstrate its effectiveness in refining noise 
predictions in the high-frequency domain. To visualize the spatial 
distribution of error improvement, we compute the high-frequency error 
difference map:
\[
\Delta E_{\mathrm{HF}}
= \left\lVert \hat{\varepsilon}_{\text{baseline}} - \varepsilon \right\rVert_{\mathrm{HF}}
- \left\lVert \hat{\varepsilon}_{\text{LRRB}} - \varepsilon \right\rVert_{\mathrm{HF}}.
\]

The high-frequency components are extracted using FFT-based high-pass 
filtering with a radial cutoff of \( r_c = 0.8 \) (top 20\% frequencies). 
To improve the visibility of small differences, we apply a sign-preserving 
logarithmic transformation:
\[
\Delta E_{\mathrm{HF}}^{(\log)}
= \operatorname{sign}(\Delta E_{\mathrm{HF}})
\cdot \log\!\left( 1 + \left| \Delta E_{\mathrm{HF}} \right| \right).
\]

\begin{figure}
    \centering
    \includegraphics[width=1\linewidth]{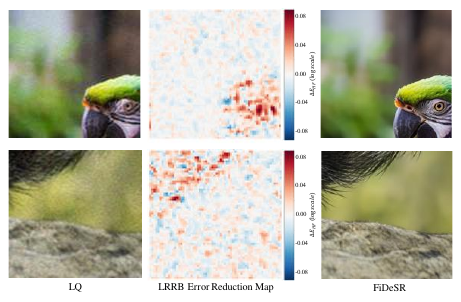}
    \caption{Spatial distribution of high-frequency noise prediction error 
improvement by LRRB. $\Delta E_{\mathrm{HF}}$ represents the 
difference between baseline and LRRB error magnitudes in the 
high-frequency domain (top 20\% frequencies, $r_c = 0.8$). 
Positive values (red) indicate regions where LRRB reduces 
prediction error, while negative values (blue) indicate regions 
where error increases. Color intensity represents the magnitude 
of change in log scale.}
    \label{fig:LRRB}
\end{figure}

Table~\ref{tab:errorreduction} in the main paper presents the quantitative 
error values, while Fig.~\ref{fig:LRRB} in this supplement visualizes these 
improvements. Positive values (red) indicate regions where LRRB reduces 
prediction error, whereas negative values (blue) correspond to areas where 
the error increases. The visualization further reveals that improvements 
(red regions) are concentrated in perceptually important areas such as 
edges and fine textures, where accurate noise prediction is essential for 
preserving image fidelity. These results demonstrate that, by incorporating 
LRRB, the FiDeSR model better preserves high-frequency details from 
low-quality inputs, enabling more realistic and visually faithful image 
restoration.

\subsection{DAW Module Visualization}

\begin{figure}
    \centering
    \includegraphics[width=1\linewidth]{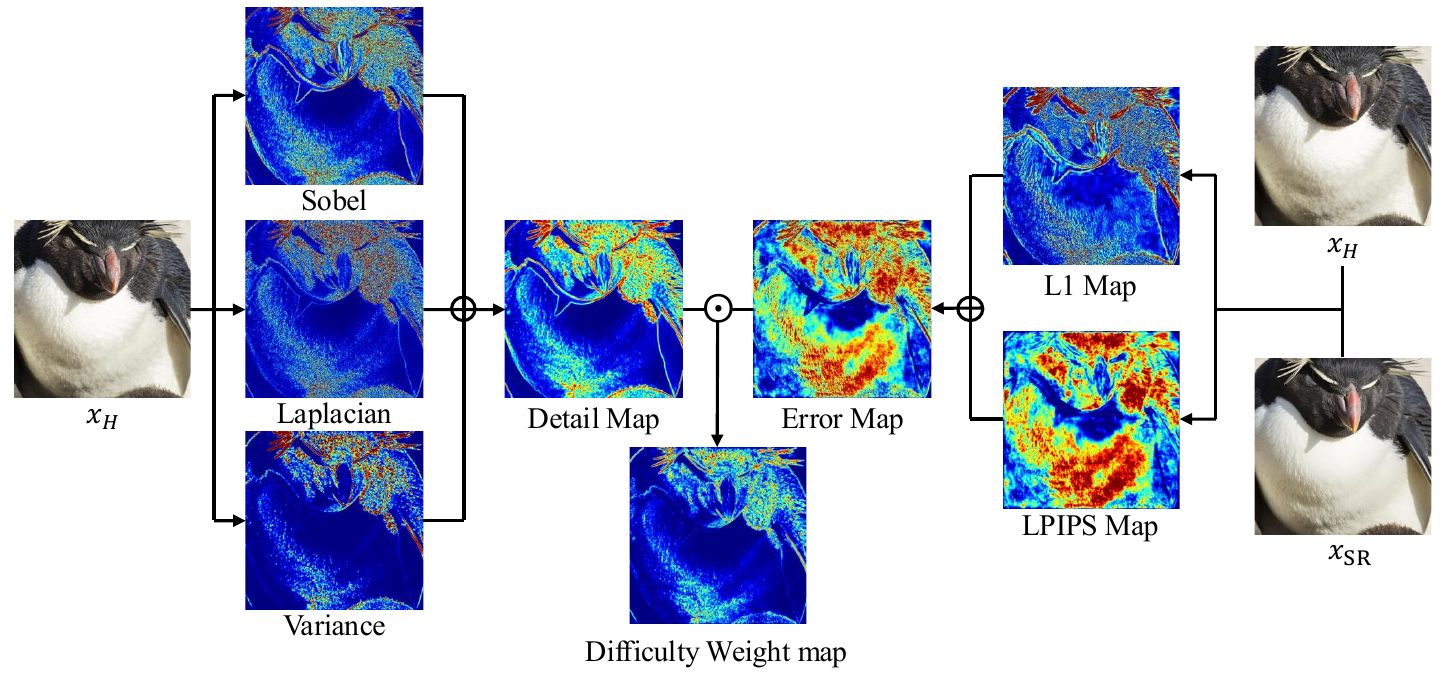}
    \caption{Visualization of the Detail-aware Weighting (DAW) module. Detail Map generated by spatial filters (Sobel, Laplacian, Variance), is element-wise multiplied by the Error Map to create the Difficulty Weight Map.}
    \label{fig:DAW}
\end{figure}

As illustrated in Fig.~\ref{fig:DAW}, the Detail-aware Weighting (DAW) module generates a Detail Map using multiple spatial filters, including Sobel, Laplacian, and local variance operators. Each filter captures different types of fine structures, enabling the model to extract complementary detail cues from the input image. The resulting Detail Map is then element-wise multiplied with the Error Map to produce the Difficulty Weight Map, which emphasizes regions where the model tends to make larger prediction errors. By guiding the network to focus more on these challenging and structurally important areas, the DAW module effectively enhances the perceptual quality of the reconstructed images.

\subsection{Qualitative Analysis of LRRB and DAW Contributions}
\begin{figure*}
    \centering
    \includegraphics[width=1\linewidth]{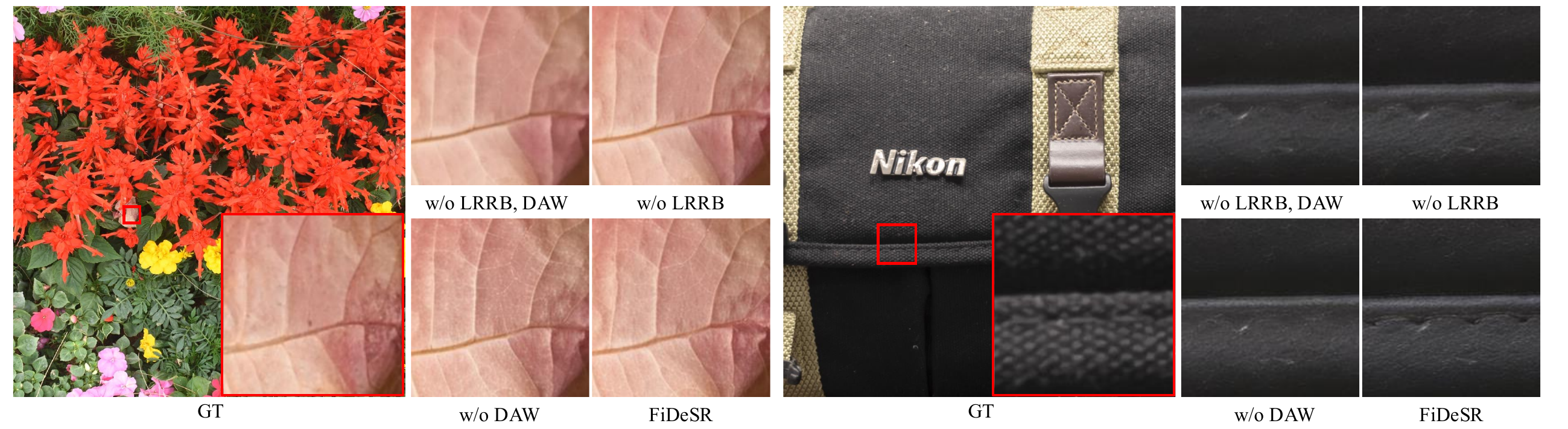}
    \caption{Qualitative ablation study illustrating the contributions of the LRRB and DAW modules.
Even under challenging degradation conditions, the full FiDeSR model reconstructs finer details
and preserves texture fidelity most effectively.}
    \label{fig:ablation}
\end{figure*}

To further analyze the effectiveness of each component in FiDeSR, we conduct a qualitative ablation study, as shown in Fig.~\ref{fig:ablation}. By individually removing the LRRB and DAW modules, we observe clear degradation in the reconstruction quality. Excluding the LRRB reduces the model's ability to refine high-frequency structures, leading to blurry textures and loss of fine details. Without the DAW module, the model becomes less sensitive to spatially challenging regions, resulting in artifacts and reduced perceptual sharpness. In contrast, the full FiDeSR model consistently reconstructs finer textures and preserves structural fidelity even under severe degradation conditions, demonstrating the complementary contributions of both modules.

\subsection{Ablations on LoRA rank}
Table~\ref{tab:lora_rank} presents the effect of varying the LoRA rank in FiDeSR. 
Although different ranks lead to slight variations across distortion-oriented and perceptual metrics, the overall performance remains stable and competitive. 
In our implementation, we adopt a LoRA rank of 8, which offers a strong balance between fidelity 
(PSNR/SSIM) and perceptual quality (NIQE, MUSIQ, MANIQA).
The results in the table also show that other ranks (4 and 16) achieve similarly competitive performance.

\begin{table*}[t]
\centering
\caption{Quantitative comparison with GAN-based Real-ISR Methods. Best results are highlighted in {\color{red}\textbf{red}}.}
\label{tab:sr_results_gan}
\resizebox{\textwidth}{!}{
\begin{tabular}{llccccccccc}
\toprule
\textbf{Dataset} & \textbf{Method} & \textbf{PSNR $\uparrow$} & \textbf{SSIM $\uparrow$} & \textbf{LPIPS $\downarrow$} & \textbf{DISTS $\downarrow$} & 
\textbf{CLIPIQA $\uparrow$} & \textbf{NIQE $\downarrow$} & \textbf{MUSIQ $\uparrow$} & \textbf{MANIQA $\uparrow$} & \textbf{FID $\downarrow$} \\
\midrule
\multirow{4}{*}{DrealSR}
& Real-ESRGAN & 28.61 & 0.8051 & 0.2819 & \textbf{\color{red}{0.2089}} & 0.4519 & 6.6896 & 54.2678 & 0.4904 & 147.68 \\
& BSRGAN      & 28.70 & 0.8028 & 0.2858 & 0.2144 & 0.5093 & 6.5387 & 57.1626 & 0.4844 & 155.59 \\
& LDL         & 28.20 & \textbf{\color{red}{0.8124}} & \textbf{\color{red}{0.2792}} & 0.2127 & 0.4475 & 7.1360 & 53.9464 & 0.4894 & 155.53 \\
& \textbf{FiDeSR} & \textbf{\color{red}{28.90}} & 0.7907 & 0.2836 & 0.2112 & \textbf{\color{red}{0.6974}} & \textbf{\color{red}{6.2014}} & \textbf{\color{red}{65.7820}} & \textbf{\color{red}{0.6239}} & \textbf{\color{red}{127.97}} \\
\midrule
\multirow{4}{*}{RealSR}
& Real-ESRGAN & 25.68 & 0.7614 & 0.2709 & 0.2060 & 0.4485 & 5.7936 & 60.3674 & 0.5505 & 135.20 \\
& BSRGAN      & \textbf{\color{red}{26.37}} & \textbf{\color{red}{0.7651}} & 0.2656 & 0.2124 & 0.5119 & 5.6361 & 63.2870 & 0.5420 & 141.30 \\
& LDL         & 25.28 & 0.7565 & 0.2750 & 0.2120 & 0.4554 & 5.9905 & 60.9277 & 0.5494 & 142.68 \\
& \textbf{FiDeSR} & 26.02 & 0.7457 & \textbf{\color{red}{0.2626}} & \textbf{\color{red}{0.1965}} & \textbf{\color{red}{0.6896}} & \textbf{\color{red}{5.3194}} & \textbf{\color{red}{69.8245}} & \textbf{\color{red}{0.6681}} & \textbf{\color{red}{109.68}} \\
\midrule
\multirow{4}{*}{DIV2K}
& Real-ESRGAN & 24.29 & \textbf{\color{red}{0.6372}} & 0.3112 & 0.2141 & 0.5277 & 4.6790 & 61.0621 & 0.5485 & 37.63 \\
& BSRGAN      & \textbf{\color{red}{24.58}} & 0.6269 & 0.3351 & 0.2275 & 0.5247 & 4.7510 & 61.1953 & 0.5041 & 44.22 \\
& LDL         & 23.83 & 0.6344 & 0.3256 & 0.2227 & 0.5179 & 4.8549 & 60.0382 & 0.5328 & 42.28 \\
& \textbf{FiDeSR} & 24.33 & 0.6250 & \textbf{\color{red}{0.2678}} & \textbf{\color{red}{0.1845}} & \textbf{\color{red}{0.6873}} & \textbf{\color{red}{4.6644}} & \textbf{\color{red}{68.8672}} & \textbf{\color{red}{0.6384}} & \textbf{\color{red}{23.30}} \\
\bottomrule
\end{tabular}
}
\end{table*}

\makeatletter
\def\heavyrulewidth{0.5pt}
\def\lightrulewidth{0.4pt}
\def\cmidrulewidth{0.4pt}
\makeatother

\begin{table*}[t]
\centering
\caption{Comparison of inference steps, runtime, and model parameters among diffusion-based SR methods.}
\label{tab:inference_step}

\fontsize{6}{7}\selectfont
\setlength{\tabcolsep}{7pt}
\renewcommand{\arraystretch}{0.5}

\resizebox{\textwidth}{!}{
\begin{tabular}{lccccccccc}
\specialrule{0.8pt}{2pt}{2pt}
\textbf{Metric} & StableSR & DiffBIR & SeeSR & PASD & AddSR & SinSR & OSEDiff & PISA-SR & FiDeSR (Ours) \\
\specialrule{0.8pt}{2pt}{2pt}
Inference Step & 200 & 50 & 50 & 20 & 4 & 1 & 1 & 1 & 1 \\
\midrule
Inference Time (s) & 7.52 & 2.04 & 3.30 & 2.10 & 0.76 & 0.097 & 0.087 & 0.057 & 0.078 \\
\midrule
\#Params (B) & 1.56 & 1.68 & 2.51 & 2.31 & 2.28 & 0.18 & 1.77 & 1.30 & 1.29 \\
\specialrule{0.8pt}{2pt}{2pt}
\end{tabular}
}
\end{table*}

\makeatletter
\def\heavyrulewidth{0.5pt}
\def\lightrulewidth{0.4pt}
\def\cmidrulewidth{0.4pt}
\makeatother

\begin{table*}[t]
\centering
\caption{Ablation study of LoRA rank for FiDeSR on the RealSR dataset (evaluated before applying LFIM).}
\label{tab:lora_rank}

\fontsize{6}{7}\selectfont
\setlength{\tabcolsep}{7pt}
\renewcommand{\arraystretch}{0.5}

\resizebox{\textwidth}{!}{
\begin{tabular}{c c c c c c c c c c}
\specialrule{0.8pt}{2pt}{2pt}

\textbf{LoRA Rank} & 
\textbf{PSNR $\uparrow$} & 
\textbf{SSIM $\uparrow$} & 
\textbf{LPIPS $\downarrow$} &
\textbf{DISTS $\downarrow$} & 
\textbf{CLIPIQA $\uparrow$} & 
\textbf{NIQE $\downarrow$} &
\textbf{MUSIQ $\uparrow$} & 
\textbf{MANIQA $\uparrow$} & 
\textbf{FID $\downarrow$} \\
\midrule
4  & 26.5196 & 0.7577 & 0.2573 & 0.1967 & 0.6554 & 5.3488 & 68.4282 & 0.6442 & 112.9637 \\
\midrule
8  & 26.2542 & 0.7498 & 0.2604 & 0.1963 & 0.6764 & 5.3332 & 69.3610 & 0.6580 & 108.7867 \\
\midrule
16 & 26.4582 & 0.7563 & 0.2508 & 0.1920 & 0.6660 & 4.6858 & 68.0145 & 0.6525 & 108.6132 \\
\specialrule{0.8pt}{2pt}{2pt}
\end{tabular}
}
\end{table*}

\begin{algorithm}[t]
\caption{Pseudo-code of Detail-Aware Weighting}
\label{alg:daw}

\begin{algorithmic}[1]                 
\renewcommand{\algorithmicrequire}{\textbf{Input:}}
\renewcommand{\algorithmicensure}{\textbf{Output:}}

\STATE \textbf{Input:} GT $y$, pred $\hat{y}$, mix $p$    
\STATE \textbf{Output:} $L_{l2}, L_{lpips}, L_{csd}$      

\STATE $y \leftarrow \text{to\_gray}(y)$
\STATE $(S,L,V) \leftarrow (\text{Sobel}(y), \text{Laplacian}(y), \text{Variance}(y))$
\STATE $D \leftarrow \text{box\_blur3x3}(\text{quantile\_norm}((S+L+V)/3))$
\STATE $E_{\text{pix}},E_{\text{perc}} \leftarrow \text{L1}(\hat{y},y),\ \text{LPIPS}(\hat{y},y)$
\STATE $E \leftarrow (1-p)\,E_{\text{pix}} + p\,E_{\text{perc}}$
\STATE $E \leftarrow \text{quantile\_norm}(E)$
\STATE $W \leftarrow \tanh(\text{blur}(D \odot E) / w_{\max}) \cdot w_{\max}$
\STATE $w^\ast \leftarrow \text{mean\_norm}(1 + \alpha \cdot W)$
\STATE $L_{l2} \leftarrow (w^\ast \cdot (\hat{y} - y)^2).mean()$
\STATE $L_{lpips} \leftarrow (\text{resize}(w^\ast)\cdot \text{LPIPS}(\hat{y}, y)).mean()$
\STATE $L_{csd} \leftarrow (\text{resize}(w^\ast)\cdot \text{CSD}(\text{latents}, \text{prompts})).mean()$
\end{algorithmic}
\end{algorithm}

\section{Implementation Details of DAW}
\label{sec:DAW_implement}
We summarize the implementation of DAW used during training in Alg.~\ref{alg:daw}. DAW computes a detail map $D$ from spatial operators (Sobel, Laplacian, and Variance) on the HQ target $x_H$, and an error map $E$ by mixing pixel-level (L1) and perceptual (LPIPS) discrepancies between $x_{SR}$ and $x_H$ with coefficient $p$. These maps are combined to form a per-pixel difficulty weight, which is applied to both the reconstruction loss and the CSD loss. 

\section{Implementation Details of LFIM}
\label{sec:LFIM}

\noindent\textbf{Additional Details of LFIM on LF and HF.}
Beyond the main description in the paper, the inference code reveals several important implementation aspects of the low-frequency injection (LFIM on LF) and high-frequency injection (LFIM on HF) mechanisms that are not explicitly discussed in the main manuscript.

\paragraph{Frequency Decomposition and Injection Intensity.}
Both variants of LFIM operate directly in the latent space $z$ using FFT-based Butterworth filtering. 
The injection intensity is governed by two global weighting parameters:
\texttt{lf\_alpha} for low-frequency reinforcement and \texttt{hf\_beta} for high-frequency enhancement.
These coefficients determine how strongly the filtered components are injected back into the latent representation.
Larger values of \texttt{lf\_alpha} result in stronger stabilization of global structures, while higher \texttt{hf\_beta} promote more pronounced enhancement of textures and edges.
As observed in Table~\ref{tab:ablation_hflf_full} of the main paper, increasing low-frequency injection monotonically improves PSNR and SSIM, whereas stronger high-frequency injection leads to higher MUSIQ and MANIQA scores.

As summarized in Table~\ref{tab:lfhf_ratio}, adjusting both 
$\texttt{lf\_alpha}$ and $\texttt{hf\_beta}$ jointly provides a flexible way 
to control the balance between structural fidelity and perceptual sharpness, 
and both the overall injection strength and the LF/HF ratio can be freely 
manipulated depending on the desired behavior. In practice, these parameters 
allow users to steer the reconstruction preference, ranging from clean and 
stable outputs to sharper results with more pronounced texture details.

In our implementation, we adopt a balanced configuration of 
$\texttt{lf\_alpha} = 0.2$ and $\texttt{hf\_beta} = 0.2$, 
which we found to yield a well-rounded compromise between global structural 
consistency and perceptual detail enhancement. This setting avoids excessively 
biasing the model toward either distortion-centric or perception-centric 
behavior, producing stable and visually coherent results across datasets.

\paragraph{LFIM on Low Frequency}
LFIM on LF extracts the low-frequency residual $\Delta_{LP}$ through a Butterworth low-pass filter and selectively injects it back into $z$ using spatial and channel gating.
The spatial gate is derived from Sobel, Laplacian, and local variance maps, limiting LF injection in detail-rich regions to avoid oversmoothing.
The channel gate evaluates Pseudo-PSD energy to identify channels dominated by structural information.
The final LF injection is applied as
\[
z \leftarrow z 
+ \texttt{lf\_alpha} 
\cdot M_{\mathrm{sp}} 
\cdot M_{\mathrm{ch}} 
\cdot \Delta_{LP},
\]
with optional morphological erosion available to refine the spatial mask boundaries.
This mechanism stabilizes illumination, coarse geometry, and tone consistency, reducing structural distortion in the restored output.

\paragraph{LFIM on High Frequency}
LFIM on HF extracts the high-frequency component $\Delta_{HP}$ either directly or using the differential high-pass term $\mathrm{HPF}(z)-\mathrm{HPF}(z_{\tilde{}})$ when enabled by \texttt{hf\_use\_diff}.
Spatial gating emphasizes edge regions through a detail-dependent exponent $\gamma$, while channel gating selects frequency-rich channels complementary to the LF gate.
The final injection is computed as
\[
z \leftarrow z 
+ \texttt{hf\_beta} 
\cdot M_{\mathrm{sp}}^{HF} 
\cdot M_{\mathrm{ch}}^{HF} 
\cdot \Delta_{HP}.
\]
This selective enhancement sharpens textures, edges, and micro-patterns without altering global luminance structure, and results in substantial boosts to perceptual metrics.

\paragraph{Summary.}
LFIM on LF enhances global structural fidelity, whereas LFIM on HF improves perceptual sharpness.  
The use of balanced injection intensities enables FiDeSR to recover both reliable structure and rich detail, ultimately yielding high-quality, frequency-aware super-resolution results.

\begin{table}[t]
\centering
\caption{Ablation study on different LF/HF injection strength ratios of LFIM $(\texttt{lf\_alpha}, \texttt{hf\_beta})$ on the RealSR dataset.}
\resizebox{0.48\textwidth}{!}{%
\begin{tabular}{lccccc}
\toprule
\textbf{LF/HF Ratio} & \textbf{PSNR} $\uparrow$ & \textbf{SSIM} $\uparrow$ 
& \textbf{CLIPIQA} $\uparrow$ & \textbf{MUSIQ} $\uparrow$ & \textbf{MANIQA} $\uparrow$ \\
\midrule
(0.2,\ 0.2) & 26.0249 & 0.7457 & 0.6896 & 69.8245 & 0.6681 \\
(0.4,\ 0.2) & 26.0658 & 0.7464 & 0.6877 & 69.7645 & 0.6666 \\
(0.4,\ 0.4) & 25.8630 & 0.7428 & 0.6934 & 70.0136 & 0.6737 \\
(0.6,\ 0.6) & 25.6519 & 0.7391 & 0.6956 & 70.1429 & 0.6789 \\
\bottomrule
\end{tabular}
}
\label{tab:lfhf_ratio}
\end{table}

\begin{table}[t]
\centering
\caption{Comparison between FiDeSR and TFDSR on SR benchmarks.}

\label{tab:fidesr_tfdsr}

\resizebox{1\linewidth}{!}{
\begin{tabular}{llccccc}
\toprule
\textbf{Dataset} & \textbf{Method} 
& \textbf{PSNR $\uparrow$} 
& \textbf{LPIPS $\downarrow$} 
& \textbf{NIQE $\downarrow$} 
& \textbf{MANIQA $\uparrow$} 
& \textbf{FID $\downarrow$} \\
\midrule
\multirow{2}{*}{DRealSR}
 & TFDSR  & 27.88 & 0.3417 & 6.2667 & 0.6164 & 155.66 \\
 & FiDeSR & \textcolor{red}{\textbf{28.90}} & \textcolor{red}{\textbf{0.2836}} & \textcolor{red}{\textbf{6.2014}} & \textcolor{red}{\textbf{0.6239}} & \textcolor{red}{\textbf{127.97}} \\
\bottomrule
\end{tabular}
}
\end{table}

\section{Comparison with Frequency-Aware Diffusion SR}
\label{sec:freq}
In Table~\ref{tab:fidesr_tfdsr}, we provide a quantitative comparison with TFDSR \cite{li2025timestep}, a frequency-aware diffusion SR method, on the DRealSR dataset. FiDeSR consistently outperforms TFDSR in both full-reference and no-reference metrics. Specifically, FiDeSR achieves a lower LPIPS score, indicating that the restored images are more perceptually and semantically consistent with the ground-truth. Furthermore, improved NIQE and MANIQA scores indicate that FiDeSR generates more naturalistic details and higher visual quality than TFDSR.